\pdfoutput=1

\documentclass[11pt]{article}

\usepackage[]{EMNLP2022}

\usepackage{times}
\usepackage{latexsym}

\usepackage[T1]{fontenc}

\usepackage[utf8]{inputenc}

\usepackage{microtype}

\usepackage{inconsolata}

\usepackage{amssymb}
\usepackage{amsmath}
\usepackage{graphicx}
\usepackage{titlesec}
\titleformat{\subsubsection}[runin]{\bfseries}{}{}{}[]
\usepackage{subcaption}
\newcommand\correspondingauthor{\thanks{*Corresponding author.}}

\title{Multi-level Distillation of Semantic Knowledge for Pre-training Multilingual Language Model}

\author{Mingqi Li$^1$, Fei Ding$^1$, Dan Zhang$^1$, Long Cheng$^1$, Hongxin Hu$^2$, Feng Luo$^1$\correspondingauthor\\
        $^1$Clemson University, $^2$University at Buffalo\\
        \texttt{\{mingqil,feid,dzhang4,lcheng2,luofeng\}@clemson.edu}\\ \texttt{hongxinh@buffalo.edu}}

\begin{document}
\maketitle
\begin{abstract}
Pre-trained multilingual language models play an important role in cross-lingual natural language understanding tasks. However, existing methods did not focus on learning the semantic structure of representation, and thus could not optimize their performance. In this paper, we propose Multi-level Multilingual Knowledge Distillation (MMKD), a novel method for improving multilingual language models. Specifically, we employ a teacher-student framework to adopt rich semantic representation knowledge in English BERT. We propose token-, word-, sentence-, and structure-level alignment objectives to encourage multiple levels of consistency between source-target pairs and correlation similarity between teacher and student models. We conduct experiments on cross-lingual evaluation benchmarks including XNLI, PAWS-X, and XQuAD. Experimental results show that MMKD outperforms other baseline models of similar size on XNLI and XQuAD and obtains comparable performance on PAWS-X. Especially, MMKD obtains significant performance gains on low-resource languages.
\end{abstract}

\section{Introduction}

Pre-training a large-scale language model and fine-tuning it on downstream tasks has shown great success in natural language processing. Most works (\citealp{devlin2018bert}; \citealp{liu2019roberta}; \citealp{yang2019xlnet}) focused on English, since it is easy to get a large amount of training data for English. This paradigm has recently emerged as a promising means for cross-lingual tasks. Some multilingual language models (\citealp{devlin2018bert}; \citealp{conneau2019unsupervised}) trained on monolingual data from over 100 languages using Masked Language Modeling (MLM) objective performed surprisingly well without any explicit alignment. On the other hand, XLM \citep{lample2019cross} extended MLM objective to a parallel corpus version - Translation Language Modeling (TLM), and achieved impressive results. This inspired researchers to develop alignment methods using parallel corpora.

Follow-up works of XLM (\citealp{yang2020alternating}; \citealp{wei2020learning}; \citealp{chi2020infoxlm}; \citealp{ouyang2020ernie}) leveraged various of training objectives to align parallel sentences at different granularity. These works were usually trained using both large amounts of monolingual data and additional parallel corpora, which are time-consuming and require considerable computational resources. 

Another line of research (\citealp{cao2020multilingual}; \citealp{pan2020multilingual}; \citealp{hu2020explicit}) only used limited parallel data to improve existing pre-trained language models rather than training new models from scratch. They depended on new alignment methods for parallel pairs of words and sentences, which could further achieve performance gains over current state-of-the-art pre-trained language models. However, these approaches neglected vector space properties when aligning across languages, and thus generating sub-optimal results. We hypothesize that a large-scale English corpus can provide more semantic and structural information than most other languages used to train multilingual language models. Moreover, BERT \citep{devlin2018bert}, which is trained from a vast amount of English Wikipedia and BooksCorpus~\citep{zhu2015aligning}, can capture this information properly and guide the training procedure of other languages, especially for those with limited resources.  

In this work, we employ a teacher-student framework to adopt vector space properties in English, and transfer its rich knowledge to our multilingual language model. We use BERT-base as the teacher model and Multilingual BERT (mBERT; \citealp{devlin2018bert}) as the student model. We propose a Multi-level Multilingual Knowledge Distillation (MMKD) method to align semantically similar sentences in parallel corpora to improve mBERT. Specifically, we propose a Cross-lingual Word-aware Contrastive Learning (XWCL) to encourage word representation similarity between teacher and student networks. We also adopt TLM objective in the student network to take advantage of corresponding context information in the target languages of masked tokens. We present a new sentence-level alignment objective to imitate English sentence projections from the teacher network. Moreover, we propose a structure-level alignment objective to transfer relationships between sentences in BERT vector space. We conduct experiments on zero-shot cross-lingual natural language understanding tasks, including natural language inference, paraphrase identification, and question answering. Experimental results show that MMKD significantly improves mBERT and outperforms baseline models of similar size. The analysis demonstrates the cross-lingual transferability of MMKD on low-resource languages. MMKD provides a more feasible and effective pre-training procedure that only requires limited training data and fewer computational resources.

\section{Related Work}

\subsection{Multilingual Language Model Pre-training}

Several efforts trained multilingual language models with transformer-based architectures and large-scale monolingual corpora across over 100 languages. For instance, \citet{devlin2018bert} trained Multilingual BERT (mBERT) on 104 languages with objectives of Masked Language Modeling (MLM) and Next Sentence Prediction (NSP), and it performed surprisingly well without any explicit alignment. XLM-R \citep{conneau2019unsupervised} achieved further performance gains by leveraging more training data and a larger model.

One popular family of methods proposed different training objectives to align words or sentences from parallel corpora. XLM \citep{lample2019cross} extended MLM objective to a parallel corpus version - Translation Language Modeling (TLM). Unicoder \citep{huang2019unicoder} improved the transferability by presenting five pre-training tasks. ALM \citep{yang2020alternating} predicted words in the context of code-switching sentences. HICTL \citep{wei2020learning} introduced sentence-level and word-level alignment with contrastive learning. INFOXLM \citep{chi2020infoxlm} proposed cross-lingual contrast (XLCO) to maximize mutual information of sentence pairs. ERNIE-M \citep{ouyang2020ernie} presented cross-attention masked language modeling (CAMLM) and back-translation masked modeling (BTMLM) to leverage both parallel and monolingual corpora.

More recently, researchers considered computational resources and time and presented works based on existing multilingual language models. \citet{cao2020multilingual} minimized the similarity between word pairs in parallel sentences in a post-hoc manner.  \citet{pan2020multilingual} argued that creating word alignments using FastAlign \citep{dyer2013simple} would suffer from the noise of the toolkit and neglected the contextual information. They proposed Post-Pretraining Alignment (PPA) that combined a different TLM objective and a contrastive learning objective. AMBER~\citep{hu2020explicit} presented objectives that encouraged prediction of the corresponding sentence and consistency between attention matrices, and they pre-trained the model with an extremely large batch size of 8,192 for the first 1M steps.

\subsection{Knowledge Distillation}

\citet{hinton2015distilling} first introduced knowledge distillation to transfer knowledge to a small model, and it has been widely used for transferring dark knowledge (which refers to information that can tell us how the model tends to generalize) and model compression in Natural Language Processing and Computer Vision. A series of follow-up works achieved gains on multilingual tasks. \citet{sun2020knowledge} enhanced the generalization ability of unsupervised neural machine translation by adding self-knowledge distillation and language branch knowledge distillation. \citet{wang2020structure} reduced the distance between monolingual teachers and the multilingual student to predict multilingual label sequences. To the best of our knowledge, \citet{reimers2020making} is the only multilingual language model related work that applied a student model to mimic sentence representations generated from the teacher model. They fed both source and target sentences into the student model to calculate Mean Square Error (MSE) loss with the teacher model's source sentences.

\begin{figure*}[h]
  \centering
    \includegraphics[width=16cm]{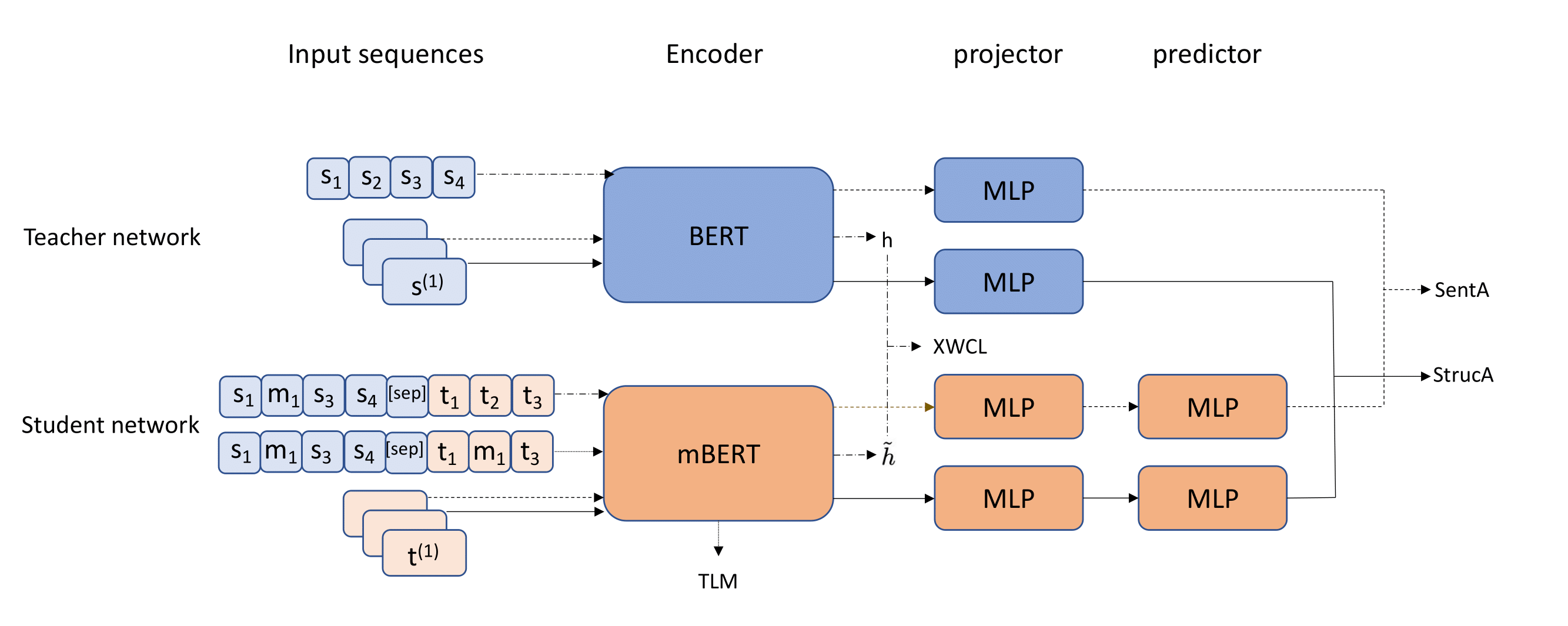}
  \caption{Model Architecture of our proposed Multi-level Multilingual Knowledge Distillation method which combines TLM, XWCL, SentA, and StrucA objectives and is trained in a multi-task manner.}\label{figure:model}
\end{figure*}

\section{Methodology}

This section presents the training procedure and introduces our four proposed training objectives. Our goal is to improve multilingual language models by transferring semantic knowledge from English and aligning multi-level information in parallel corpora with limited resources. The general network architecture is illustrated in Figure~\ref{figure:model}. The student network consists of three components: an encoder, a projector, and a predictor, while the teacher network contains an encoder and a projector. 


\subsection{Translation Language Modeling}

Translation Language Modeling (TLM) objective is an extension of MLM~\citep{lample2019cross}. Given the concatenation of parallel sentences, TLM objective predicts masks in both source and target sequences. In this way, TLM utilizes context information in the corresponding language, and thus helps the model to learn token-level alignments.

Similar to~\citet{devlin2018bert}, we randomly mask 15\% tokens from input sequences and replace them with a [MASK] token 80\% of the time, with a random token in vocabulary 10\% of the time, and keep them unchanged 10\% of the time. The input sequence is denoted as $[s_1,\ldots,s_a,$ [SEP]$, t_1,\ldots,t_b]$, where a, b are numbers of tokens, and masks exist in both source and target sides. Since the teacher model only involves English, we train TLM objective on the student model.

\subsection{Cross-lingual Word-aware Contrastive Learning}

Inspired by~\citet{su2021tacl}, we propose a cross-lingual version of word-aware contrastive learning (XWCL) objective. The goal of XWCL is to encourage the student model to learn more discriminative representations. Different from~\citet{su2021tacl}, our student model produces representations according to the parallel context instead of surrounding monolingual words. Moreover, due to the vocabulary difference in our teacher and student models, we align the representations on the word-level.

Given an English source sequence $s=[s_1,\ldots,s_n]$ and a target sequence $t=[t_1,\ldots,t_m]$, we concatenate them with a special token [SEP] and randomly mask 15\% words only from source sequence $s$ following the same mask strategy in~\citet{devlin2018bert}. Then, we feed this masked sequence into the student model and get representation $\tilde{h}$ = [$\tilde{h_1},\ldots,\tilde{h}_{n+m}$]. Meanwhile, we input the original sequence $s$ into the teacher model and get $h=[h_1,\ldots,h_n]$ as reference. Our proposed XWCL objective learns to minimize the infoNCE loss of the masked tokens:
\begin{equation}
\resizebox{0.15\hsize}{!}{$\mathcal{L}_{\mathrm{XWCL}}=-$}
\resizebox{0.25\hsize}{!}{$\sum_{i=1}^{n} \log m \left(s_{i}\right)$}
\resizebox{0.45\hsize}{!}{$\frac{\exp \left(\operatorname{sim}\left(\tilde{h}_{i}, h_{i}\right) / \tau\right)}{\sum_{j=1}^{n} \exp \left(\operatorname{sim}\left(\tilde{h}_{i}, h_{j}\right) / \tau\right)},$}
\end{equation}
where $\tau$ is a temperature parameter, sim(·,·) denotes dot product, $m \left(s_{i}\right)=1$ if $s_{i}$ is a masked token, otherwise $m \left(s_{i}\right)=0$. Here we mask the whole word and treat the first token of each mask as the word representation. Consequently, XWCL will make masked representations produced by the student model closer to their corresponding representations in English vector space, and discriminate them from other distinct representations. 

\subsection{Sentence Alignment}

BERT is well-trained with a large-scale English corpus and thus encodes rich semantic knowledge. The goal of our proposed Sentence Alignment (SentA) objective is to capture this semantic information and transfer it to mBERT. Similar to \citet{grill2020bootstrap}, we learn representations by instance-level discrimination without negative samples, while we freeze the teacher model rather than updating with an exponential moving average.

Given a sentence pair $(s^{(i)},t^{(i)})$ in parallel corpora, where $s^{(i)}$ is the i-th sentence from English and $t^{(i)}$ is from a target language, we treat them as two different views and input $s^{(i)}$ into the teacher network and $t^{(i)}$ into the student network separately. We minimize Mean Squared Error loss between teacher projections and student predictions:
\begin{equation}
\resizebox{0.89\hsize}{!}{$
\mathcal{L}_{\mathrm{SentA}}=
\frac{1}{|\mathcal{B}|} \sum_{i \in \mathcal{B}}\left(\bar{q}_{\theta}\left(g_{\theta}\left(\tilde{c}^{(i)}\right)\right)-\bar{g}_{\xi}\left(c^{(i)}\right)\right)^{2},$}
\end{equation}
where $c^{(i)}$ and $\tilde{c}^{(i)}$ are the [CLS] tokens of last hidden states of i-th sentence in the teacher and student encoders, $g$ defines the projectors with distinct parameters and $q$ defines the predictor, and $\bar{g}$, $\bar{q}$ indicate that they are normalized with $L_2$ norm. More precisely, we apply 2-layers MLPs to implement projectors and the predictor. 

SentA objective will force different languages closer to semantically similar English sentences in the vector space. Meanwhile, the student network can adopt well-trained English vector space properties by imitating corresponding representations in the teacher network.

\subsection{Structure Alignment}

Transferring relationships between samples plays a crucial role in knowledge distillation. Inspired by~\citet{ding2020multi}, we propose a Structure Alignment (StrucA) objective to learn knowledge correlation.

Given a batch of source-target sentence pairs $((s^{(1)},t^{(1)}),\ldots,(s^{(\mathcal{B})},t^{(\mathcal{B})}))$, we feed them into the same teacher-student encoders as calculating SentA objective, while using their own projection and prediction heads. Let $z = [z^{(1)},\ldots,z^{(\mathcal{B})}]$ and $\tilde{z}$ = [$\tilde{z}^{(1)},\ldots,\tilde{z}^{(\mathcal{B})}$] denote teacher projections and student predictions. The proposed objective allows the student network to mimic the vector space structure of the teacher network, which means the correlation between $\tilde{z}$ is similar to $z$. Specifically, we first normalize $z$ and calculate the similarity matrix:
\begin{equation}
\mathcal{A}_{i, j}=z^{(i)} \cdot z^{(j)}, \tilde{\mathcal{A}}_{i, j}=\tilde{z}^{(i)} \cdot \tilde{z}^{(j)}.
\end{equation}
Then, the teacher's relational function can be expressed as:
\begin{equation}
\resizebox{0.3\hsize}{!}{$\psi\left(z^{(1)}, . ., z^{(\mathcal{B})}\right)$}
=\frac{\exp \left(\mathcal{A}_{i, j} / \tau\right)}{\sum_{j} \exp \left(\mathcal{A}_{i, j} / \tau\right)}.
\end{equation}
The student network follows the same step, but takes $log\_softmax$ function as a relational function instead. Finally, we employ KL-divergence loss to minimize the difference between two probability distributions:
\begin{equation}
\mathcal{L}_{\text {StrucA}}=\sum_{i=1}^{\mathcal{B}} \text{KLDivLoss} \left(\psi\left(\cdot \right),\tilde{\psi}\left(\cdot \right)\right).
\end{equation}


After training, the relationship between samples produced by the student network in vector space will be similar to their counterparts in English. In conclusion, StrucA objective learns additional structural information in English vector space.

\subsection{Multi-level Multilingual Knowledge Distillation Pre-training}

We jointly train these proposed objectives that cover alignments at different granularity, and the final loss would be:
\begin{equation}
\mathcal{L}=\mathcal{L}_{\mathrm{TLM}}+\mathcal{L}_{\mathrm{XWCL}}+\mathcal{L}_{\mathrm{SentA}}+\alpha\mathcal{L}_{\mathrm{StrucA}},
\end{equation}
where $\alpha$ is used to balance the weights.

For training, we update the student network by AdamW~\citep{loshchilov2017decoupled} optimizer, while freezing parameters in the teacher model.

In addition, we randomly shuffle the sentence pairs from each parallel datasets, but balance the number of samples from each language within a batch. In other words, our model will consider each language of the same weight during the training procedure.

\section{Experiments}

This section explains our training details and shows the experimental results on XNLI, PAWS-X and XQuAD. We compare our proposed MMKD with existing works following the setting in \citet{hu2020xtreme} and conduct ablation studies to prove the effectiveness of each proposed objective.

\begin{table*}
\centering
\resizebox{1.75\columnwidth}{!}{\begin{tabular}{cccccccc}
\hline
\textbf{Model} & \textbf{Vocab size} & \textbf{Layers} & \textbf{Parameters} & \textbf{Ratio} & \textbf{Data}\\
\hline
\text{mBERT} & \text{119K} & \text{12} & \text{178M} & \text{1.0x} & \text{Wikipedia}\\
\text{MONOTRANS} & \text{30K} & \text{12} & \text{110M} & \text{0.62x} & \text{Wikipedia}\\
\text{PPA} & \text{110K} & \text{12} & \text{172M} & \text{0.97x} & \text{10.5M parallel data}\\
\text{AMBER} & \text{120K} & \text{12} & \text{172M} & \text{0.97x} & \text{Wikipedia + 58.5M parallel data}\\
\text{\citet{cao2020multilingual}} & \text{119K} & \text{12} & \text{178M} & \text{1.0x} & \text{1.8M parallel data}\\
\textbf{MMKD (this work)} & \text{119K} & \text{12} & \text{179M} & \text{1.01x} & \text{10.5M parallel data}\\
\text{MMTE} & \text{64K} & \text{6} & \text{192M} & \text{1.08x} & \text{103 languages in-house parallel data}\\
\text{mT5} & \text{250K} & \text{12} & \text{580M} & \text{3.26x} & \text{CommonCrawl}\\
\text{XLM-100} & \text{200K} & \text{12} & \text{828M} & \text{4.65x} & \text{Wikipedia}\\
\text{XLM-R-Large} & \text{250K} & \text{24} & \text{816M} & \text{4.58x} & \text{CommonCrawl}\\
\hline
\end{tabular}}
\caption{Model size and training data for comparison. Ratio is the parameters' ratio of mBERT. Wikipedia and CommonCrawl are extremely larger than other parallel datasets. For AMBER, we only list the parallel data size of the languages we consider. The numbers of parameters in PPA and AMBER are slightly different from mBERT we use.}
\label{tab:modelarch}
\end{table*}

\subsection{Training Details}

\subsubsection*{Training Data}
\hspace{0.25cm}We collect the same parallel corpora as previous works (\citealp{cao2020multilingual}; \citealp{pan2020multilingual}) for comparison. Specifically, we treat English as source language and download datasets from the OPUS website~\citep{tiedemann2012parallel} including (1) low-resource languages: en-hi from IITB~\citep{kunchukuttan2017iit} and en-bg from EUbookshop and Europarl~\citep{koehn2005europarl} (2) high-resource languages: en-ar, en-zh from MultiUN~\citep{eisele2010multiun} and en-fr, en-es, en-de from Europarl. Our training data does not involve any monolingual corpora; however, mBERT is trained from large-scale monolingual corpora. Additionally, we remove extremely short (less than 10 tokens) and long (more than 128 tokens) sentences and prune each dataset to 2M sentence pairs if they contain more than that. Table~\ref{tab:modelarch} indicates the size of parallel datasets in our work.

\subsubsection*{Model Architecture}
\hspace{0.25cm}The architecture of teacher and student encoders is the same as BERT-base, which contains 12 layers, 768 hidden states, and 12 attention heads. The student encoder is initialized from mBERT, while the teacher encoder is initialized from BERT-base, and thus they have different vocabulary sizes. Additionally, the student encoder is followed by two different projection and prediction heads, and two projection heads are also on the top of the teacher encoder. These heads consist of randomly initialized 2-layer MLP with 768 hidden dimensions and 128 output dimensions.

\subsubsection*{Training Setups}
\hspace{0.25cm}During the training procedure, we optimize the student network by AdamW with 1e-2 weight decay and schedule the learning rate with a linear decay peaking at 2e-5 after 10\% warm-up steps. We set 128 tokens as the maximum length of each sequence and use a batch size of 256. The training procedure takes 3 days for 15 epochs on 8 40GB Nvidia A100 GPUs. For the evaluation procedure, we fine-tune the student encoder for few epochs with a batch size of 32 on English training data, and evaluate on target languages.

\subsection{Evaluation Benchmarks}

We evaluate our multilingual language model using publicly available cross-lingual natural language understanding benchmarks, including natural language inference, paraphrase identification, and question answering tasks. We conduct all the experiments with a zero-shot setting: we fine-tune the model on English training data and directly test on target languages.

\subsubsection*{XNLI}
\hspace{0.25cm}\citet{conneau2018xnli} is a widely used cross-lingual sentence classification dataset that extends SNLI/MultiNLI (\citealp{bowman2015large}; \citealp{N18-1101}) in fifteen languages. The task is to classify the relationships of two given sentences to entailment, neutral, or contradiction. This dataset provides 2490 dev samples and 5010 test samples in each language. In the zero-shot setting, we fine-tune our student encoder using English MultiNLI, which contains 392,702 sentence pairs. Then, we select the model according to the performance on XNLI English dev set, and test target languages using XNLI test set.

\subsubsection*{PAWS-X}
\hspace{0.25cm} The goal of PAWS-X~\citep{yang2019paws} is to identify whether the two sentences are paraphrases. This dataset translates Paraphrase Adversaries from Word Scrambling (PAWS)~\citep{zhang2019paws} evaluation pairs in six languages. We take 49,401 English training pairs in PAWS as training data, and use around 2,000 sentence pairs of each target language from PAWS-X as testing data.

\begin{table*}
\centering
\resizebox{1.85\columnwidth}{!}{\begin{tabular}{l|cccccccc|c}
\hline
\text{Models} & \text{en} & \text{ar} & \text{bg} & \text{de} & \text{es} & \text{fr} & \text{hi} & \text{zh} & \text{avg}\\
\hline
\multicolumn{1}{c}{\text{Model size similar to mBERT}}\\
\hline
\text{mBERT~\citep{devlin2018bert}} & \text{81.4} & \text{62.1} & \text{-} & \text{70.5} & \text{74.3} & \text{-} & \text{-} & \text{63.8} & \text{-}\\
\text{mBERT*~\citep{devlin2018bert}} & \text{80.8} & \text{64.3} & \text{68.0} & \text{70.0} & \text{73.5} & \text{73.4} & \text{58.9} & \text{67.8} & \text{69.6}\\
\text{MONOTRANS~\citep{artetxe2019cross}} & \text{81.7} & \text{70.6} & \text{73.7} & \text{73.0} & \text{75.4} & \text{74.7} & \text{65.2} & \text{70.3} & \text{73.1}\\
\text{\citet{cao2020multilingual}} & \text{80.1} & \text{-} & \text{73.4} & \text{73.1} & \text{75.5} & \text{74.5} & \text{-} & \text{-} & \text{-}\\
\text{PPA~\citep{pan2020multilingual}} & \text{82.8} & \text{70.3} & \text{73.8} & \text{74.2} & \text{76.7} & \text{76.6} & \text{66.9} & \textbf{72.8} & \text{74.3}\\
\text{AMBER~\citep{hu2020explicit}} & \textbf{84.7} & \text{70.2} & \text{74.3} & \text{74.2} & \text{76.9} & \text{76.6} & \text{66.2} & \text{71.6} & \text{74.3}\\
\text{MMTE*~\citep{siddhant2020evaluating}} & \text{79.6} & \text{64.9} & \text{70.4} & \text{68.2} & \text{71.6} & \text{69.5} & \text{63.5} & \text{69.2} & \text{69.6}\\
\textbf{MMKD (this work)} & \text{83.2} & \textbf{71.6} & \textbf{75.8} & \textbf{76.2} & \textbf{78.0} & \textbf{77.2} & \textbf{69.0} & \text{72.3} & \textbf{75.4}\\
\hline
\multicolumn{1}{l}{\text{Larger models}}\\
\hline
\text{mT5-Base~\citep{xue2020mt5}} & \text{84.7} & \text{73.3} & \text{78.6} & \text{77.4} & \text{80.3} & \text{79.1} & \text{70.8} & \text{74.1} & \text{77.3}\\
\text{XLM-100*~\citep{lample2019cross}} & \text{82.8} & \text{66.0} & \text{71.9} & \text{72.7} & \text{75.5} & \text{74.3} & \text{62.5} & \text{70.2} & \text{72.0}\\
\text{XLM-R-Large*~\citep{conneau2019unsupervised}} & \textbf{88.7} & \textbf{77.2} & \textbf{83.0} & \textbf{82.5} & \textbf{83.7} & \textbf{82.2} & \textbf{75.6} & \textbf{78.2} & \textbf{81.4}\\
\hline
\end{tabular}}
\caption{Zero-shot cross-lingual classification evaluation results on XNLI. * indicates the results are taken from~\citet{hu2020xtreme}. All other results are from original papers.}
\label{tab:xnli}
\end{table*}

\subsubsection*{XQuAD}
\hspace{0.25cm}\citet{artetxe2019cross} requires to return an answer span derived from the paragraph according to the question. Professional translators translate a subset of SQuAD v1.1~\citep{rajpurkar2016squad} development set into ten languages, containing 1,290 question-answering pairs. We use 87,599 training data in SQuAD v1.1 together with 1,190 testing data of each target language in XQuAD.

\subsection{Results and Analysis}
We report the results across the above evaluation benchmarks. We compare our pre-trained model with mBERT~\citep{devlin2018bert} and the following mBERT-based models: (1)~\citet{cao2020multilingual}; (2) PPA~\citep{pan2020multilingual}; (3) AMBER~\citep{hu2020explicit}. These models adopt BERT-base architecture and are initialized from mBERT. We also compare with MONOTRANS~\citep{artetxe2019cross} and MMTE~\citep{siddhant2020evaluating}, which contain a similar amount of parameters to mBERT. These models use relatively fewer computational resources than larger models, but still achieve some gains. We also take the results of large models as reference including: (1) mT5~\citep{xue2020mt5}; (2) XLM-100~\citep{lample2019cross}; (3) XLM-R-Large~\citep{conneau2019unsupervised}. These models are usually costly, but they boost the state-of-the-art on many cross-lingual tasks. There is a resources-performance trade-off in training multilingual language models. Table~\ref{tab:modelarch} shows the model size and training data.

\subsubsection*{XNLI}
\hspace{0.25cm}Table~\ref{tab:xnli} presents zero-shot cross-lingual classification accuracy on XNLI. Similar to~\citet{cao2020multilingual} and~\citet{pan2020multilingual}, we evaluate only on the languages used in the pre-training procedure. We first compare with zero-shot results of mBERT to see whether our alignment method improves the existing model. We take the mBERT results from ~\citet{hu2020xtreme} to provide a more comprehensive comparison. Our model significantly outperforms mBERT across all the reported languages. We obtain a boost in performance of 5.8\% accuracy on average. Moreover, we observe significant improvements on Bulgarian and Hindi which are considered as low-resource languages in our experiments. We only collect 370k en-bg sentence pairs and 895k en-hi sentence pairs. However, they outperform mBERT by 7.8\% and 10.1\% respectively. 

Compared to the models of similar size, we reach a state-of-the-art of 75.4\% across eight languages on the zero-shot XNLI benchmark dataset. We significantly outperform these models on ar, bg, de, es, fr, hi and achieve comparable results on en and zh. The results show that we obtain consistent improvements except Chinese compared to PPA~\citep{pan2020multilingual}. Additionally, the performance of Latin-based languages is better than non-Latin-based languages in our case. We conclude that our method is more beneficial to languages closed to English, since we adopt English BERT as the teacher model and all the training sentence pairs involve English.

Our method also produces 3.4\% gains compared to XLM-100, despite the fact that we have 78\% fewer parameters than theirs. Our model is 6\% less than XLM-R-Large, which is 4.5 times larger and employs a much more extensive training dataset.

\begin{table*}
\centering
\resizebox{1.4\columnwidth}{!}{\begin{tabular}{l|ccccc|c}
\hline
\text{Models} & \text{en} & \text{de} & \text{es} & \text{fr} & \text{zh} & \text{avg}\\
\hline
\multicolumn{1}{l}{\text{Model size similar to mBERT}}\\
\hline
\text{mBERT*~\citep{devlin2018bert}} & \text{94.0} & \text{85.7} & \text{87.4} & \text{87.0} & \text{77.0} & \text{86.2}\\
\text{MONOTRANS~\citep{artetxe2019cross}} & \text{94.3} & \text{86.3} & \text{87.6} & \text{87.3} & \text{79.0} & \text{86.9}\\
\text{AMBER~\citep{hu2020explicit}} & \textbf{95.6} & \textbf{89.4} & \text{89.2} & \textbf{90.7} & \text{80.9} & \textbf{89.2}\\
\text{MMTE*~\citep{siddhant2020evaluating}} & \text{93.1} & \text{85.1} & \text{87.2} & \text{86.9} & \text{75.9} & \text{85.6}\\
\textbf{MMKD (this work)} & \text{94.7} & \text{88.8} & \textbf{89.7} & \text{88.9} & \textbf{81.1} & \text{88.6}\\
\hline
\multicolumn{1}{l}{\text{Larger models}}\\
\hline
\text{mT5-Base~\citep{xue2020mt5}} & \textbf{95.4} & \text{89.4} & \text{89.6} & \textbf{91.2} & \text{81.1} & \text{89.2}\\
\text{XLM-100*~\citep{lample2019cross}} & \text{94.0} & \text{85.9} & \text{88.3} & \text{87.4} & \text{76.5} & \text{86.4}\\
\text{XLM-R-Large*~\citep{conneau2019unsupervised}} & \text{94.7} & \textbf{89.7} & \textbf{90.1} & \text{90.4} & \textbf{82.3} & \textbf{89.4}\\
\hline
\end{tabular}}
\caption{Zero-shot cross-lingual paraphrase identification evaluation accuracy on PAWS-X. * indicates the results are taken from~\citet{hu2020xtreme}. All other results are from original papers. }
\label{tab:pawsx}
\end{table*}

\begin{table*}
\centering
\resizebox{1.5\columnwidth}{!}{\begin{tabular}{l|cccccc|c}
\hline
\text{Models} & \text{en} & \text{ar} & \text{de} & \text{es} & \text{hi} & \text{zh} & \text{avg}\\
\hline
\multicolumn{1}{l}{\text{Model size similar to mBERT}}\\
\hline
\text{mBERT*~\citep{devlin2018bert}} & \text{83.5} & \text{61.5} & \text{70.6} & \text{75.5} & \text{59.2} & \text{58.0} & \text{68.1}\\
\text{MONOTRANS~\citep{artetxe2019cross}} & \text{82.1} & \textbf{66.0} & \text{70.6} & \text{70.8} & \text{61.9} & \textbf{60.5} & \text{68.7}\\
\text{MMTE*~\citep{siddhant2020evaluating}} & \text{80.1} & \text{63.2} & \text{68.8} & \text{72.4} & \text{61.3} & \text{55.8} & \text{66.9}\\
\textbf{MMKD (this work)} & \textbf{84.8} & \text{64.0} & \textbf{73.5} & \textbf{76.7} & \textbf{62.7} & \text{58.8} & \textbf{70.1}\\
\hline
\multicolumn{1}{l}{\text{Larger models}}\\
\hline
\text{mT5-Base~\citep{xue2020mt5}} & \text{84.6} & \text{63.8} & \text{73.8} & \text{74.8} & \text{60.3} & \textbf{66.1} & \text{70.6}\\
\text{XLM-100*~\citep{lample2019cross}} & \text{74.2} & \text{61.4} & \text{66.0} & \text{68.2} & \text{56.6} & \text{49.7} & \text{62.7}\\
\text{XLM-R*~\citep{conneau2019unsupervised}} & \textbf{86.5} & \textbf{68.6} & \textbf{80.4} & \textbf{82.0} & \textbf{76.7} & \text{59.3} & \textbf{75.6}\\
\hline
\end{tabular}}
\caption{Zero-shot cross-lingual question answering evaluation F1 score on XQuAD. * indicates the results are taken from~\citet{hu2020xtreme}. All other results are from original papers.}
\label{tab:xquad}
\end{table*}

\subsubsection*{PAWS-X}
\hspace{0.25cm}Table~\ref{tab:pawsx} reports zero-shot cross-lingual paraphrase identification accuracy on PAWS-X. In this experiment, we evaluate on five high-resource languages involved in our pre-training step. The high resource helps to learn rich information and structured semantic representations; thus, existing models can perform well across these languages. We push mBERT classification accuracy from 86.2\% to 88.6\% with the help of alignment objectives.

MMKD outperforms models of similar size by an accuracy of 1.7\%-3\% except AMBER. One primary reason is that AMBER is trained with an extremely large batch size that has proven effective by \citet{liu2019roberta}. Another reason is that PAWS-X only consists of high-resource languages, and thus can not demonstrate our model's benefit on low-resource languages. Similar to XNLI, we observe consistent improvements over XLM-100 on the paraphrase identification task. Compared to XLM-R-Large, we bridge the performance gap to 0.8\% with limited computational resources. 

\subsubsection*{XQuAD}
\hspace{0.25cm}Table~\ref{tab:xquad} shows zero-shot cross-lingual question answering F1 score on XQuAD. Our model obtains the best F1 score on average against other baseline models of similar size. The results on four Latin-based languages significantly outperform other models, while our model produces relatively small gains on Arabic and Chinese. This is consistent with our findings on XNLI.

For larger models, we outperform XLM-100 across all evaluation languages and achieve comparable results to mT5-Base whose parameters are much more than ours.

\subsection{Ablation Study}

We conduct ablation studies to investigate the impact of each objective in our framework.

\begin{figure*}
\centering
\begin{subfigure}{.5\textwidth}
  \centering
  \includegraphics[width=\linewidth]{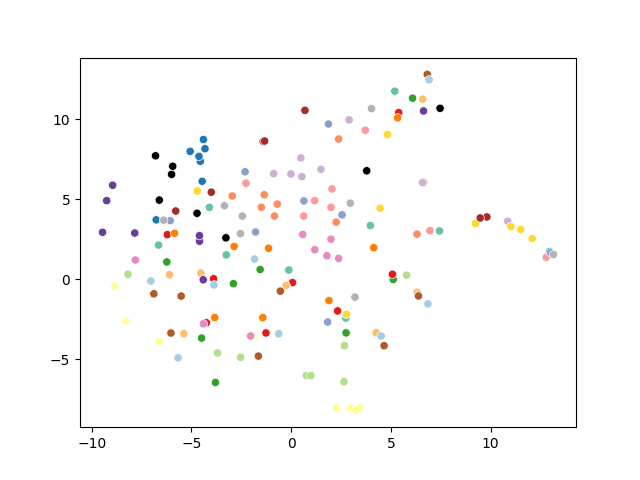}
  \caption{representations from mBERT}
  \label{fig:mbert}
\end{subfigure}%
\begin{subfigure}{.5\textwidth}
  \centering
  \includegraphics[width=\linewidth]{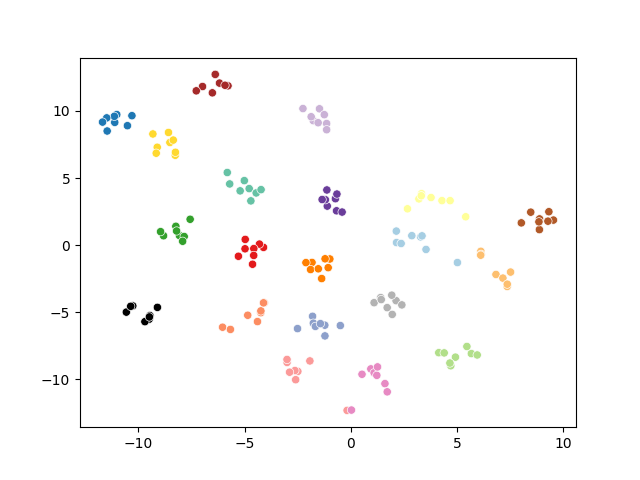}
  \caption{representations from MMKD}
  \label{fig:mmkd}
\end{subfigure}
\caption{tSNE plots of 20 sentence representations on XNLI 15-way parallel corpus. We take eight languages used in our pre-training procedure. Dots in the same color are the representation of English and representation of its translations into other seven languages.}
\label{fig:tsne}
\end{figure*}

In this experiment, we remove each training objective respectively from our original model and get four pre-trained multilingual language models. Compared to the original MMKD, we can measure the performance gains of each training objective. 

We report the average results across languages we evaluated on benchmark datasets in Table~\ref{tab:ablation}. We employ identical training setups to minimize the effect of other factors. We observe performance drops on ablated models across all the evaluation benchmarks.

On XNLI benchmark, MMKD outperforms other ablated models by 1.3\%-2.0\%. This result demonstrates that removing either proposed alignment objective will lead to less semantic knowledge. 
We can observe that performance drops dramatically on PAWS-X benchmark without the structure-level training objective. This indicates that aligning knowledge correlation between teacher and student models can benefit obtaining semantic information to distinguish sentences with similar words. 

Similar to findings for XNLI and PAWS-X, F1 score on XQuAD benchmark worsens by 0.4\% to 8.9\% without each training objective. Sentence-level alignment has a great impact on this question answering task.

In conclusion, each proposed training objective provides various semantic and structure knowledge and contributes to performance improvement.

\begin{table}
\centering
\resizebox{0.7\columnwidth}{!}{\begin{tabular}{l|ccc}
\hline
\text{Models} & \text{XNLI} & \text{PAWS-X} & \text{XQuAD}\\
\hline
\text{Metrics} & \text{Acc} & \text{Acc} & \text{F1}\\
\hline
\text{MMKD} & \textbf{75.4} & \textbf{88.6} & \textbf{70.1}\\
\text{-XWCL} & \text{74.1} & \text{86.4} & \text{69.7}\\
\text{-TLM} & \text{73.5} & \text{87.5} & \text{68.5}\\
\text{-SentA} & \text{73.4} & \text{86.4} & \text{61.2}\\
\text{-StrucA} & \text{73.9} & \text{85.7} & \text{69.2}\\
\hline
\end{tabular}}
\caption{Zero-shot cross-lingual results on evaluation benchmarks. MMKD indicates that the original model was pre-trained with all proposed objectives. - indicates the training objective which is removed from our framework, and thus the model is pre-trained using the other three training objectives.}
\label{tab:ablation}
\end{table}
\subsection{Visualization of Representations}
To further assess the effectiveness of our proposed alignment method, we visualize the sentence representations of MMKD and original mBERT using t-SNE~\citep{van2008visualizing}. We utilize a 15-way corpus provided by XNLI~\citep{conneau2018xnli}. This corpus contains 10,000 sentences and their translations in fifteen languages. We randomly select 20 sentences and their translations in seven languages used in the pre-training procedure.

Figure~\ref{fig:tsne} shows t-SNE plots of these sentence representations. Each dot represents a sentence representation produced by the multilingual language model. We treat [CLS] token embedding of last hidden states as the sentence representation. The dots in the same color are translations from the same sentence; thus, we have 8 dots in each color. Figure~\ref{fig:mbert} and Figure~\ref{fig:mmkd} show representation t-SNE projections from mBERT and MMKD respectively. We observe that semantically similar sentences from different languages are clustered in vector space by MMKD, while these representations from mBERT do not follow this trend.

This visualization result confirms that our alignment method makes semantically similar sentences closed in the vector space even though they are from different languages. This result also proves the effectiveness of our method for transferring semantic knowledge from English to other languages.

\section{Conclusion}
In this work, we propose a Multi-level Multilingual Knowledge Distillation method to pre-train the multilingual language model - mBERT. We propose four training objectives to align token-, word-, and sentence-level information from parallel corpora, and we also learn knowledge correlation between teacher and student models. Compared to existing studies, we require fewer computational resources and less training time. In the zero-shot cross-lingual setting, our model outperforms models of similar size and reduces the performance gap to larger models across XNLI, PAWS-X, and XQuAD benchmarks. Experimental results show that MMKD obtains significant performance gains on low-resource languages and does well on Latin-based languages. Visualization result shows that semantic relationships among sentences have been successfully transferred from English to other languages. Future work could extend our approach to other larger multilingual language models.

\section*{Limitations}
In order to adopt rich vector space properties, we utilize English BERT as our teacher model during the pre-training procedure. MMKD achieves impressive results on Indo-European languages that are closed to English, while performance on languages from other language families that are distantly related to English get less improved. For example, Arabic and Chinese are high-resource languages whose performance across three evaluation tasks is lower than those of Indo-European languages. However, they still have performance gains compared to models of same size. Future work could consider combining multiple teacher models covering various language families.

\section*{Ethics Statement}
The authors of this work follow the ACL Code of Ethics. This work complies with the ACL Ethics Policy.

\section*{Acknowledgements}
The work was supported in part by the U.S. National Science Foundation (NSF) under Grant MRI-2018069 and Grant SES-2031002 to Feng Luo, Long Cheng, and Hongxin Hu.

\bibliography{anthology,custom}

\begin{thebibliography}{38}
\expandafter\ifx\csname natexlab\endcsname\relax\def\natexlab#1{#1}\fi

\bibitem[{Artetxe et~al.(2019)Artetxe, Ruder, and Yogatama}]{artetxe2019cross}
Mikel Artetxe, Sebastian Ruder, and Dani Yogatama. 2019.
\newblock On the cross-lingual transferability of monolingual representations.
\newblock \emph{arXiv preprint arXiv:1910.11856}.

\bibitem[{Bowman et~al.(2015)Bowman, Angeli, Potts, and
  Manning}]{bowman2015large}
Samuel~R Bowman, Gabor Angeli, Christopher Potts, and Christopher~D Manning.
  2015.
\newblock A large annotated corpus for learning natural language inference.
\newblock \emph{arXiv preprint arXiv:1508.05326}.

\bibitem[{Cao et~al.(2020)Cao, Kitaev, and Klein}]{cao2020multilingual}
Steven Cao, Nikita Kitaev, and Dan Klein. 2020.
\newblock Multilingual alignment of contextual word representations.
\newblock \emph{arXiv preprint arXiv:2002.03518}.

\bibitem[{Chi et~al.(2020)Chi, Dong, Wei, Yang, Singhal, Wang, Song, Mao,
  Huang, and Zhou}]{chi2020infoxlm}
Zewen Chi, Li~Dong, Furu Wei, Nan Yang, Saksham Singhal, Wenhui Wang, Xia Song,
  Xian-Ling Mao, Heyan Huang, and Ming Zhou. 2020.
\newblock Infoxlm: An information-theoretic framework for cross-lingual
  language model pre-training.
\newblock \emph{arXiv preprint arXiv:2007.07834}.

\bibitem[{Conneau et~al.(2019)Conneau, Khandelwal, Goyal, Chaudhary, Wenzek,
  Guzm{\'a}n, Grave, Ott, Zettlemoyer, and Stoyanov}]{conneau2019unsupervised}
Alexis Conneau, Kartikay Khandelwal, Naman Goyal, Vishrav Chaudhary, Guillaume
  Wenzek, Francisco Guzm{\'a}n, Edouard Grave, Myle Ott, Luke Zettlemoyer, and
  Veselin Stoyanov. 2019.
\newblock Unsupervised cross-lingual representation learning at scale.
\newblock \emph{arXiv preprint arXiv:1911.02116}.

\bibitem[{Conneau et~al.(2018)Conneau, Rinott, Lample, Williams, Bowman,
  Schwenk, and Stoyanov}]{conneau2018xnli}
Alexis Conneau, Ruty Rinott, Guillaume Lample, Adina Williams, Samuel~R.
  Bowman, Holger Schwenk, and Veselin Stoyanov. 2018.
\newblock Xnli: Evaluating cross-lingual sentence representations.
\newblock In \emph{Proceedings of the 2018 Conference on Empirical Methods in
  Natural Language Processing}. Association for Computational Linguistics.

\bibitem[{Devlin et~al.(2018)Devlin, Chang, Lee, and
  Toutanova}]{devlin2018bert}
Jacob Devlin, Ming-Wei Chang, Kenton Lee, and Kristina Toutanova. 2018.
\newblock Bert: Pre-training of deep bidirectional transformers for language
  understanding.
\newblock \emph{arXiv preprint arXiv:1810.04805}.

\bibitem[{Ding et~al.(2020)Ding, Yang, Hu, Krovi, and Luo}]{ding2020multi}
Fei Ding, Yin Yang, Hongxin Hu, Venkat Krovi, and Feng Luo. 2020.
\newblock Multi-level knowledge distillation via knowledge alignment and
  correlation.
\newblock \emph{arXiv preprint arXiv:2012.00573}.

\bibitem[{Dyer et~al.(2013)Dyer, Chahuneau, and Smith}]{dyer2013simple}
Chris Dyer, Victor Chahuneau, and Noah~A Smith. 2013.
\newblock A simple, fast, and effective reparameterization of ibm model 2.
\newblock In \emph{Proceedings of the 2013 Conference of the North American
  Chapter of the Association for Computational Linguistics: Human Language
  Technologies}, pages 644--648.

\bibitem[{Eisele and Chen(2010)}]{eisele2010multiun}
Andreas Eisele and Yu~Chen. 2010.
\newblock Multiun: A multilingual corpus from united nation documents.
\newblock In \emph{LREC}.

\bibitem[{Grill et~al.(2020)Grill, Strub, Altch{\'e}, Tallec, Richemond,
  Buchatskaya, Doersch, Avila~Pires, Guo, Gheshlaghi~Azar
  et~al.}]{grill2020bootstrap}
Jean-Bastien Grill, Florian Strub, Florent Altch{\'e}, Corentin Tallec, Pierre
  Richemond, Elena Buchatskaya, Carl Doersch, Bernardo Avila~Pires, Zhaohan
  Guo, Mohammad Gheshlaghi~Azar, et~al. 2020.
\newblock Bootstrap your own latent-a new approach to self-supervised learning.
\newblock \emph{Advances in Neural Information Processing Systems},
  33:21271--21284.

\bibitem[{Hinton et~al.(2015)Hinton, Vinyals, Dean
  et~al.}]{hinton2015distilling}
Geoffrey Hinton, Oriol Vinyals, Jeff Dean, et~al. 2015.
\newblock Distilling the knowledge in a neural network.
\newblock \emph{arXiv preprint arXiv:1503.02531}, 2(7).

\bibitem[{Hu et~al.(2020{\natexlab{a}})Hu, Johnson, Firat, Siddhant, and
  Neubig}]{hu2020explicit}
Junjie Hu, Melvin Johnson, Orhan Firat, Aditya Siddhant, and Graham Neubig.
  2020{\natexlab{a}}.
\newblock Explicit alignment objectives for multilingual bidirectional
  encoders.
\newblock \emph{arXiv preprint arXiv:2010.07972}.

\bibitem[{Hu et~al.(2020{\natexlab{b}})Hu, Ruder, Siddhant, Neubig, Firat, and
  Johnson}]{hu2020xtreme}
Junjie Hu, Sebastian Ruder, Aditya Siddhant, Graham Neubig, Orhan Firat, and
  Melvin Johnson. 2020{\natexlab{b}}.
\newblock Xtreme: A massively multilingual multi-task benchmark for evaluating
  cross-lingual generalisation.
\newblock In \emph{International Conference on Machine Learning}, pages
  4411--4421. PMLR.

\bibitem[{Huang et~al.(2019)Huang, Liang, Duan, Gong, Shou, Jiang, and
  Zhou}]{huang2019unicoder}
Haoyang Huang, Yaobo Liang, Nan Duan, Ming Gong, Linjun Shou, Daxin Jiang, and
  Ming Zhou. 2019.
\newblock Unicoder: A universal language encoder by pre-training with multiple
  cross-lingual tasks.
\newblock \emph{arXiv preprint arXiv:1909.00964}.

\bibitem[{Koehn(2005)}]{koehn2005europarl}
Philipp Koehn. 2005.
\newblock Europarl: A parallel corpus for statistical machine translation.
\newblock In \emph{Proceedings of machine translation summit x: papers}, pages
  79--86.

\bibitem[{Kunchukuttan et~al.(2017)Kunchukuttan, Mehta, and
  Bhattacharyya}]{kunchukuttan2017iit}
Anoop Kunchukuttan, Pratik Mehta, and Pushpak Bhattacharyya. 2017.
\newblock The iit bombay english-hindi parallel corpus.
\newblock \emph{arXiv preprint arXiv:1710.02855}.

\bibitem[{Lample and Conneau(2019)}]{lample2019cross}
Guillaume Lample and Alexis Conneau. 2019.
\newblock Cross-lingual language model pretraining.
\newblock \emph{arXiv preprint arXiv:1901.07291}.

\bibitem[{Liu et~al.(2019)Liu, Ott, Goyal, Du, Joshi, Chen, Levy, Lewis,
  Zettlemoyer, and Stoyanov}]{liu2019roberta}
Yinhan Liu, Myle Ott, Naman Goyal, Jingfei Du, Mandar Joshi, Danqi Chen, Omer
  Levy, Mike Lewis, Luke Zettlemoyer, and Veselin Stoyanov. 2019.
\newblock Roberta: A robustly optimized bert pretraining approach.
\newblock \emph{arXiv preprint arXiv:1907.11692}.

\bibitem[{Loshchilov and Hutter(2017)}]{loshchilov2017decoupled}
Ilya Loshchilov and Frank Hutter. 2017.
\newblock Decoupled weight decay regularization.
\newblock \emph{arXiv preprint arXiv:1711.05101}.

\bibitem[{Ouyang et~al.(2020)Ouyang, Wang, Pang, Sun, Tian, Wu, and
  Wang}]{ouyang2020ernie}
Xuan Ouyang, Shuohuan Wang, Chao Pang, Yu~Sun, Hao Tian, Hua Wu, and Haifeng
  Wang. 2020.
\newblock Ernie-m: enhanced multilingual representation by aligning
  cross-lingual semantics with monolingual corpora.
\newblock \emph{arXiv preprint arXiv:2012.15674}.

\bibitem[{Pan et~al.(2020)Pan, Hang, Qi, Shah, Potdar, and
  Yu}]{pan2020multilingual}
Lin Pan, Chung-Wei Hang, Haode Qi, Abhishek Shah, Saloni Potdar, and Mo~Yu.
  2020.
\newblock Multilingual bert post-pretraining alignment.
\newblock \emph{arXiv preprint arXiv:2010.12547}.

\bibitem[{Rajpurkar et~al.(2016)Rajpurkar, Zhang, Lopyrev, and
  Liang}]{rajpurkar2016squad}
Pranav Rajpurkar, Jian Zhang, Konstantin Lopyrev, and Percy Liang. 2016.
\newblock Squad: 100,000+ questions for machine comprehension of text.
\newblock \emph{arXiv preprint arXiv:1606.05250}.

\bibitem[{Reimers and Gurevych(2020)}]{reimers2020making}
Nils Reimers and Iryna Gurevych. 2020.
\newblock Making monolingual sentence embeddings multilingual using knowledge
  distillation.
\newblock \emph{arXiv preprint arXiv:2004.09813}.

\bibitem[{Siddhant et~al.(2020)Siddhant, Johnson, Tsai, Ari, Riesa, Bapna,
  Firat, and Raman}]{siddhant2020evaluating}
Aditya Siddhant, Melvin Johnson, Henry Tsai, Naveen Ari, Jason Riesa, Ankur
  Bapna, Orhan Firat, and Karthik Raman. 2020.
\newblock Evaluating the cross-lingual effectiveness of massively multilingual
  neural machine translation.
\newblock In \emph{Proceedings of the AAAI conference on artificial
  intelligence}, volume~34, pages 8854--8861.

\bibitem[{Su et~al.(2021)Su, Liu, Meng, Lan, Shu, Shareghi, and
  Collier}]{su2021tacl}
Yixuan Su, Fangyu Liu, Zaiqiao Meng, Tian Lan, Lei Shu, Ehsan Shareghi, and
  Nigel Collier. 2021.
\newblock Tacl: Improving bert pre-training with token-aware contrastive
  learning.
\newblock \emph{arXiv preprint arXiv:2111.04198}.

\bibitem[{Sun et~al.(2020)Sun, Wang, Chen, Utiyama, Sumita, and
  Zhao}]{sun2020knowledge}
Haipeng Sun, Rui Wang, Kehai Chen, Masao Utiyama, Eiichiro Sumita, and Tiejun
  Zhao. 2020.
\newblock Knowledge distillation for multilingual unsupervised neural machine
  translation.
\newblock \emph{arXiv preprint arXiv:2004.10171}.

\bibitem[{Tiedemann(2012)}]{tiedemann2012parallel}
J{\"o}rg Tiedemann. 2012.
\newblock Parallel data, tools and interfaces in opus.
\newblock In \emph{Lrec}, volume 2012, pages 2214--2218. Citeseer.

\bibitem[{Van~der Maaten and Hinton(2008)}]{van2008visualizing}
Laurens Van~der Maaten and Geoffrey Hinton. 2008.
\newblock Visualizing data using t-sne.
\newblock \emph{Journal of machine learning research}, 9(11).

\bibitem[{Wang et~al.(2020)Wang, Jiang, Bach, Wang, Huang, and
  Tu}]{wang2020structure}
Xinyu Wang, Yong Jiang, Nguyen Bach, Tao Wang, Fei Huang, and Kewei Tu. 2020.
\newblock Structure-level knowledge distillation for multilingual sequence
  labeling.
\newblock \emph{arXiv preprint arXiv:2004.03846}.

\bibitem[{Wei et~al.(2020)Wei, Weng, Hu, Xing, Yu, and Luo}]{wei2020learning}
Xiangpeng Wei, Rongxiang Weng, Yue Hu, Luxi Xing, Heng Yu, and Weihua Luo.
  2020.
\newblock On learning universal representations across languages.
\newblock \emph{arXiv preprint arXiv:2007.15960}.

\bibitem[{Williams et~al.(2018)Williams, Nangia, and Bowman}]{N18-1101}
Adina Williams, Nikita Nangia, and Samuel Bowman. 2018.
\newblock A broad-coverage challenge corpus for sentence understanding through
  inference.
\newblock In \emph{Proceedings of the 2018 Conference of the North American
  Chapter of the Association for Computational Linguistics: Human Language
  Technologies, Volume 1 (Long Papers)}, pages 1112--1122. Association for
  Computational Linguistics.

\bibitem[{Xue et~al.(2020)Xue, Constant, Roberts, Kale, Al-Rfou, Siddhant,
  Barua, and Raffel}]{xue2020mt5}
Linting Xue, Noah Constant, Adam Roberts, Mihir Kale, Rami Al-Rfou, Aditya
  Siddhant, Aditya Barua, and Colin Raffel. 2020.
\newblock mt5: A massively multilingual pre-trained text-to-text transformer.
\newblock \emph{arXiv preprint arXiv:2010.11934}.

\bibitem[{Yang et~al.(2020)Yang, Ma, Zhang, Wu, Li, and
  Zhou}]{yang2020alternating}
Jian Yang, Shuming Ma, Dongdong Zhang, Shuangzhi Wu, Zhoujun Li, and Ming Zhou.
  2020.
\newblock Alternating language modeling for cross-lingual pre-training.
\newblock In \emph{Proceedings of the AAAI Conference on Artificial
  Intelligence}, volume~34, pages 9386--9393.

\bibitem[{Yang et~al.(2019{\natexlab{a}})Yang, Zhang, Tar, and
  Baldridge}]{yang2019paws}
Yinfei Yang, Yuan Zhang, Chris Tar, and Jason Baldridge. 2019{\natexlab{a}}.
\newblock Paws-x: A cross-lingual adversarial dataset for paraphrase
  identification.
\newblock \emph{arXiv preprint arXiv:1908.11828}.

\bibitem[{Yang et~al.(2019{\natexlab{b}})Yang, Dai, Yang, Carbonell,
  Salakhutdinov, and Le}]{yang2019xlnet}
Zhilin Yang, Zihang Dai, Yiming Yang, Jaime Carbonell, Russ~R Salakhutdinov,
  and Quoc~V Le. 2019{\natexlab{b}}.
\newblock Xlnet: Generalized autoregressive pretraining for language
  understanding.
\newblock \emph{Advances in neural information processing systems}, 32.

\bibitem[{Zhang et~al.(2019)Zhang, Baldridge, and He}]{zhang2019paws}
Yuan Zhang, Jason Baldridge, and Luheng He. 2019.
\newblock Paws: Paraphrase adversaries from word scrambling.
\newblock \emph{arXiv preprint arXiv:1904.01130}.

\bibitem[{Zhu et~al.(2015)Zhu, Kiros, Zemel, Salakhutdinov, Urtasun, Torralba,
  and Fidler}]{zhu2015aligning}
Yukun Zhu, Ryan Kiros, Rich Zemel, Ruslan Salakhutdinov, Raquel Urtasun,
  Antonio Torralba, and Sanja Fidler. 2015.
\newblock Aligning books and movies: Towards story-like visual explanations by
  watching movies and reading books.
\newblock In \emph{Proceedings of the IEEE international conference on computer
  vision}, pages 19--27.

\end{thebibliography}
\bibliographystyle{acl_natbib}




\end{document}